\pdfoutput=1

\documentclass[11pt]{article}

\usepackage[final]{coling}

\usepackage{times}
\usepackage{latexsym}

\usepackage[T1]{fontenc}

\usepackage[utf8]{inputenc}

\usepackage{microtype}

\usepackage{inconsolata}

\usepackage{color}
\definecolor{myhighlight}{RGB}{138,43,226}
\definecolor{comment}{RGB}{64,224,208}

\usepackage{graphicx}

\usepackage{amsmath,amssymb,amsfonts}%
\usepackage{amsthm}%
\usepackage{mathrsfs}%

\usepackage{verbatim} 
\usepackage{multirow}
\usepackage{booktabs} 
\usepackage{makecell}
\usepackage{CJK}
\usepackage{graphicx} 
\usepackage{caption} 
\usepackage{subcaption} 
\usepackage{setspace}
\usepackage{cleveref}
\usepackage{enumitem}

\title{Retrieval Augmented Instruction Tuning for Open NER\\ with Large Language Models}

\author{
	Tingyu Xie$^{1\mathscr{*}}$, Jian Zhang$^{1*}$, Yan Zhang$^{2}$\thanks{ ~ Equally contributed.}, \textbf{Yuanyuan Liang$^3$}, \textbf{Qi Li$^1$, Hongwei Wang$^1$}\thanks{ ~ Corresponding authors.}\\
	$^1$Zhejiang University, $^2$National University of Singapore, $^3$East China Normal University\\
	\{tingyuxie, jianzhang.22, hongweiwang\}@zju.edu.cn, yanzhang.jlu@gmail.com
}

\begin{document}
	\maketitle
	\begin{abstract}

		
		The strong capability of large language models (LLMs) has been applied to information extraction (IE) through either retrieval augmented prompting or instruction tuning (IT). However, the best way to incorporate information with LLMs for IE remains an open question. In this paper, we explore \textbf{R}etrieval \textbf{A}ugmented \textbf{I}nstruction \textbf{T}uning (\textbf{RA-IT}) for IE, focusing on the task of open named entity recognition (NER). Specifically, for each training sample, we retrieve semantically similar examples from the training dataset as the context and prepend them to the input of the original instruction. To evaluate our RA-IT approach more thoroughly, we construct a Chinese IT dataset for open NER and evaluate RA-IT in both English and Chinese scenarios. Experimental results verify the effectiveness of RA-IT across various data sizes and in both English and Chinese scenarios. We also conduct thorough studies to explore the impacts of various retrieval strategies in the proposed RA-IT framework.\footnote{Code and data are available at: \url{https://github.com/Emma1066/Retrieval-Augmented-IT-OpenNER}}

		%
		%
		%
		%
		%
		
	\end{abstract}
	
	\section{Introduction}

	The powerful generalizability of large language models (LLMs) \citep{gpt4o_blog,touvron2023llama,bai2023qwen} has been widely applied to information extraction (IE) \citep{sainz2024gollie,wang2023instructuie}. The major two lines of works for generative IE with LLMs, are prompt designing with retrieval augmented generation (RAG) using an off-the-shelf LLM \citep{wang2023gptner,guo2023retrievalaugmented,xie2024selfimproving}, and task-specific instruction tuning (IT) \citep{zhou2024universalner,sainz2024gollie,li2024knowcoder}. However, the best approach to incorporate information to LLMs for IE remains an open question. Inspired by recent studies on retrieval aware and context-enhanced IT \citep{jiang-etal-2023-active,luo2023sail,zhang2024raft,asai2024selfrag,lin2024radit,liu2024chatqa} for enhancing the LLM capability in downstream tasks, we conduct an empirical study of exploring \textbf{R}etrieval \textbf{A}ugmented \textbf{IT} (\textbf{RA-IT}) for IE, with a focus on the of open NER task.

	The previous work UniNER \citep{zhou2024universalner} distills the strong capability of ChatGPT in open NER into smaller models through IT without any human-annotated data. We follow this line and investigate RA-IT under this \textit{targeted distillation} setting. Other works of IT for IE like \citet{sainz2024gollie,li2024knowcoder} using code-style instruction data, are \textit{orthogonal} to this work since RA-IT can be integrated into various instruction styles.

	In our RA-IT approach, for each training sample, we retrieve\textit{ semantically similar examples} from the training dataset and prepend them to the original instruction, forming the context-enhanced instruction. We also explore the impacts of diverse retrieval strategies. Moreover, we construct a Chinese IT dataset for open NER and evaluate our method in both English and Chinese scenarios. We conduct thorough experiments across various data sizes and obtain the following key findings: (1) RA-IT achieves consistent improvements on various data sizes, suggesting the need for context-enhanced fine-tuning. (2) Retrieving semantically similar examples benefits the most for training among various retrieval strategies. Random retrieval also exhibits improvement but shows inferior performance to similar examples. (3) Retrieving out-domain examples for inference requires applying example filtering strategies to achieve improvements. Providing in-domain examples benefits inference.

	Our main contributions are two folds: (1) We empirically study the RA-IT framework for open NER. We prepare the retrieval augmented instruction data with semantically similar examples. We conduct thorough experimental analysis to study the impact of various retrieval strategies. (2) We construct an IT dataset for Chinese open NER and conduct our investigations in English and Chinese scenarios across various data sizes. Experimental results verify the benefits of RA-IT for open NER.

	\section{Method}

	\noindent
	\textbf{Preliminary: Targeted Distillation}.\quad
	We follow UniNER \citep{zhou2024universalner} to conduct our study in the setting of targeted distillation, where they successfully distill the strong capability of ChatGPT in open NER into smaller models, without any human-annotated data.
	The pipeline is as follows: (1) Data construction. They sample inputs from a large corpus across diverse domains, then use ChatGPT to automatically generate NER outputs. (2) Distillation. After obtaining the automatically constructed data, they apply IT to distill the open NER capability of ChatGPT into smaller models.
	
\begin{figure}[t]
		\centerline{\includegraphics[width=0.85\linewidth]{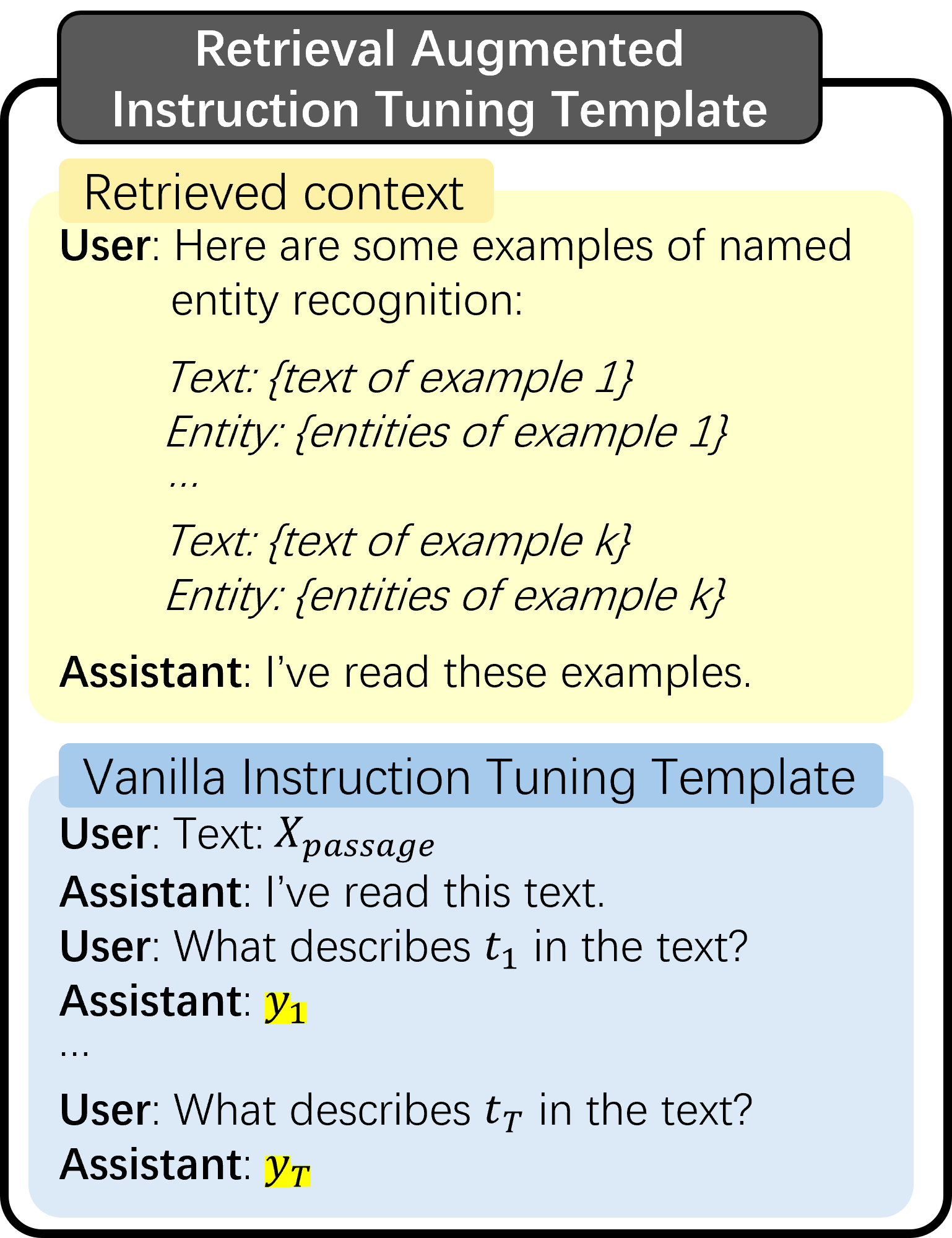}}
		\caption{The \textbf{RA-IT template}, where the \textbf{retrieved context} consists of \textit{semantically similar examples} retrieved from the training dataset and is prepended to the original \textbf{vanilla IT template}. The vanilla IT template, presented by \citet{zhou2024universalner} converts each NER sample into a conversation, where $X_{passage}$ is the input text, $[t1,\dots,t_T]$ are entity types to extract, and $y_i$ is the list of entity mentions that are $t_i$. The highlighted parts are used to compute the loss during training.}
		\label{fig:method}
		\vspace{-0.5em}
	\end{figure}

	\begin{spacing}{1.5}
	\end{spacing}
	\noindent
	\textbf{Vanilla IT}.\quad
	The original instruction tuning template used in targeted distillation is shown in the bottom part of Fig. \ref{fig:method}, which we refer to as Vanilla IT, where each passage and its associated entity output are converted into a multi-turn conversation. 

	\begin{spacing}{1.5}
	\end{spacing}
	\noindent
	\textbf{RA-IT}.\quad
	We explore an alternative way to conduct IT in targeted distillation: we introduce RA-IT, a context-enhanced tuning approach, of which the overview is in Fig. \ref{fig:method}. In our RA-IT approach, each data is augmented with a retrieved context, which consists of $k$ \textit{semantically similar examples} retrieved from the training dataset. The retrieved context is prepended to the original conversation, forming the retrieval augmented instruction. By fine tuning LMs in this recipe, we equip the LMs with the ability to generate NER answer with on-demand RAG. This means we could flexibly adapting LMs to different scenarios by determining whether to use RAG during inference based on the specific characteristics of the scenario.
	
	
	
	\begin{spacing}{1.5}
	\end{spacing}
	\noindent
	\textbf{Retriever.}\quad We use sentence embedding-based retrieval and adopt cosine similarity as our similarity metric. We retrieve the $k$ nearest neighbors as context. We also investigate various retrieval strategies for both training and inference stages.

	\begin{table*}[t]
		\centering
		\small
		\renewcommand\arraystretch{0.8}
		\begin{tabular}{c l cccccccc}
			\toprule
			Data size & Method & Movie & Restaurant & AI & Literature & Music & Politics & Science & Avg. \\
			\midrule
			- & ChatGPT & 5.30  & 32.80  & 52.40  & 39.80  & 66.60& 68.50 & 67.00  & 47.50  \\
			\midrule
			\multirow{2}{*}{5K} & Vanilla IT & 44.87  & 42.72  & 52.87  & 59.00  & 60.47  & 59.35  & 58.36  & 53.95  \\ 
			~ & RA-IT & 50.26  & 45.75  & 52.61  & 60.01  & 63.04  & 60.02  & 58.91  & \textbf{55.80}  \\
			\midrule
			\multirow{2}{*}{10K} & Vanilla IT & 49.81  & 41.47  & 53.78  & 60.99  & 63.79  & 60.84  & 61.47  & 56.02  \\ 
			~ & RA-IT & 53.79  & 45.73  & 55.90  & 62.58  & 66.52  & 62.40  & 63.67  & \textbf{58.65}  \\
			\midrule 
			\multirow{2}{*}{50K} & Vanilla IT & 44.83  & 40.39  & 58.63  & 62.88  & 64.12  & 61.63  & 63.22  & 56.53  \\ 
			~ & RA-IT & 45.18  & 40.78  & 58.01  & 63.60  & 64.76  & 61.90  & 62.79  & \textbf{56.72} \\
			\bottomrule
		\end{tabular}
		\caption{Zero-shot evaluation in English scenario. We report F1 values (\%). Numbers in \textbf{bold} indicates the best results of each category. RA-IT shows consistent improvements across various dats sizes, suggesting the need of context-enhanced training.}
		\label{tab:main_pile}
		\vspace{-0.5em}
	\end{table*}
	
	\begin{table*}[t]
		\centering
		\small
		\renewcommand\arraystretch{0.8}
		\begin{tabular}{c l ccccccccc}
			\toprule
			Data size & Method & Ontonotes 4 & MSRA & Weibo & Boson & ClueNER & CMeEE & Ren. & Yidu & Avg. \\ 
			\midrule
			- & ChatGPT & 29.70  & 41.36  & 30.25  & 46.65  & 44.75  & 43.16  & 34.25  & 34.90  & 38.13  \\
			\midrule 
			\multirow{2}{*}{5K} & Vanilla IT & 48.88  & 51.47  & 38.95  & 52.47  & 43.54  & 41.50  & 47.51  & 47.23  & 46.44  \\ 
			~ & RA-IT & 49.23  & 53.08  & 37.43  & 52.64  & 43.27  & 43.87  & 48.31  & 48.47  & \textbf{47.04}  \\ 
			\midrule
			\multirow{2}{*}{10K} & Vanilla IT & 46.28  & 52.56  & 39.26  & 52.92  & 45.42  & 42.59  & 47.99  & 47.95  & 46.87  \\ 
			~ & RA-IT & 47.69  & 55.06  & 37.38  & 53.86  & 45.25  & 43.71  & 49.25  & 47.86  & \textbf{47.51}  \\
			\midrule
			\multirow{2}{*}{50K} & Vanilla IT & 43.99  & 50.02  & 34.55  & 54.98  & 43.59  & 42.52  & 49.37  & 49.63  & 46.08  \\ 
			~ & RA-IT & 46.72  & 54.15  & 33.28  & 54.43  & 43.86  & 43.78  & 49.50  & 50.24  & \textbf{47.00} \\ \bottomrule
		\end{tabular}
		\caption{Zero-shot evaluation in Chinese scenario. We report F1 values (\%). Numbers in \textbf{bold} indicates the best results of each category. RA-IT shows consistent improvements across diverse data sizes in Chinese scenario, which further verifies the benefits of our RA-IT approach.}
		\label{tab:main_sky}
	\end{table*}
	
	\begin{CJK}{UTF8}{gbsn}
		\begin{table}[t]
			\centering
			\small
			\renewcommand\arraystretch{0.6}
			\tabcolsep=0.2em
			\begin{tabular}{ll}
				\toprule
				Frequency & Entity Type \\
				\midrule
				Top 1\% & 概念(concept), 地点(location), 人物(person),   \\
				(75.3\%) & 组织(organization), 产品(product)... \\
				\midrule
				1\%-10\% & 荣誉(honor), 技术类(technical), 场所(place),  \\
				(17.5\%)& 情绪(emotion), 节目(program)... \\
				\midrule
				10\%-100\%  & 比赛组别(competition category),   \\
				(7.2\%) &  房产类型(property type)... \\
				\bottomrule
			\end{tabular}
			\caption{Statistics of Sky-NER, the constructed IT dataset for Chinese open NER. Example entity types from various frequency ranges - top 1\%, 1-10\% and 10-100\%, along with the percentage of total frequencies for each range.}
			\label{tab:zh_it_data_stat}
			\vspace{-0.5em}
		\end{table}
	\end{CJK}

\section{Experiment}

\subsection{Experimental Settings}
\textbf{Backbones:  }We adopt LLaMA-3-8B \citep{llama3_blog} and Qwen-1.5-7B \citep{qwen1.5} as the backbone models for English and Chinese scenarios respectively. 
\textbf{Training: } For English, we use the training data Pile-NER released by \citet{zhou2024universalner}. For Chinese, we use the training data Sky-NER constructed in this paper as described in Section \ref{sec:zh_data_construction}. We use LoRA \citep{hu2021lora} to train models. Our training infrastructure was 1 NVIDIA A100 80GB.
\textbf{Retrieval: } We adopt GTE-large\footnote{https://huggingface.co/thenlper/gte-large} \citep{li2023towards} to generate text embeddings and set $k=2$ in main experiments.
\textbf{Evaluation: } We mainly focus on the zero-shot evaluation. 
For English, we adopt benchmarks CrossNER, MIT-Movie and MIT-restaurant following \citet{zhou2024universalner}. For Chinese, we collect eight benchmarks across diverse domains, of which details are in Appendix \ref{appendix:exp_setting}. We report micro-F1 value.

\subsection{Chinese IT Data Construction}
\label{sec:zh_data_construction}
Following the data construction recipe of UniNER \citep{zhou2024universalner}, we construct an IT dataset for Chinese open NER. We sample input passages from the large-scale Sky corpus \citep{wei2023skywork} across various domains, then use ChatGPT (gpt-3.5-turbo) to generate entity mentions and types based on the sampled passages. More details of data construction procedures are in Appendix \ref{sec:appendix_zh_data_construction_prompt}. We name this dataset as Sky-NER, which consists of 50K NER examples, and the type statistics are in Table \ref{tab:zh_it_data_stat}.

\begin{figure}[t]
	\centerline{\includegraphics[trim=0 0 0 0, clip,width=\linewidth]{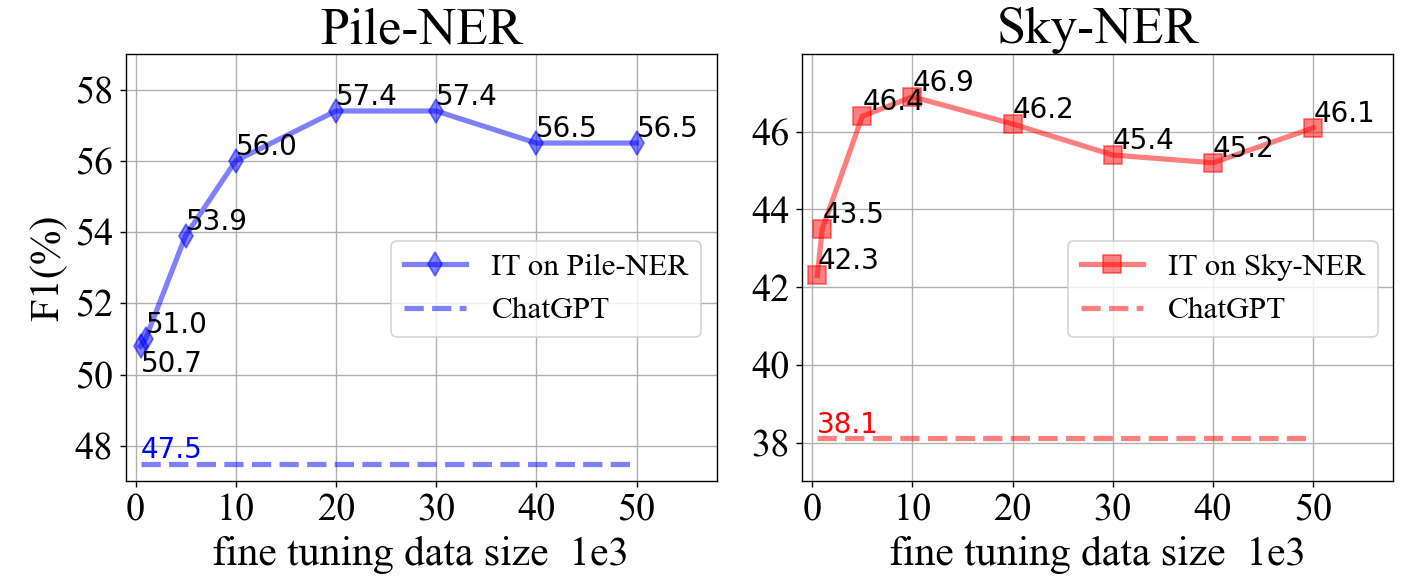}}
	\caption{Preliminary study of IT data efficiency for open NER in English (left) and Chinese (right) scenarios, where the training data are Pile-NER and Sky-NER respectively.  Average zero-shot results of evaluated benchmarks are illustrated. The performance does not necessarily improve as the data increases.}
	\label{fig:datasize}
	\vspace{-0.5em}
\end{figure}

\subsection{Preliminary Study on Data Efficiency}

We conduct a preliminary study on IT data efficiency in targeted distillation for open NER by exploring the impact of varous datas sizes: [0.5K, 1K, 5K, 10K, 20K, 30K, 40K, 50K]. We use vanilla IT for preliminary study. Results are visualized in Fig. \ref{fig:datasize}. The following observations are consistent in English and Chinese: (1) a small data size already surpass ChatGPT's performances. (2) Performances are improving as the data sizes increased to 10K or 20K, but begin to decline and then remain at a certain level as data sizes further increased to 50K. Recent work for IT data selection, \citet{xia2024less,ge2024clustering,du2023mods} also find the superior performances of only limited data size. We leave selecting more beneficial IT data for IE as future work. Accordingly, we conduct main experiments on 5K, 10K and 50K data sizes.


\subsection{Main results}
\label{sec:main_results}
The main results are summarized in Table \ref{tab:main_pile} and \ref{tab:main_sky} respectively. We report the results of inference without examples for RA-IT here, since we found this setting exhibits more consistent improvements. The impacts of inference with examples are studied in Section \ref{sec:analysis}. As shown in the tables, RA-IT shows consistent improvements on English and Chinese  across various data sizes. This presumably because the retrieved context enhance the model ability to understand the inputs. This suggests the need for context-enhanced instructions.

	\subsection{Analysis}
	\label{sec:analysis}
	We explore the impacts of diverse retrieval strategies. We conduct analysis on 5K data size for cost saving as the effect of RA-IT is consistent across various data sizes as shown in Section \ref{sec:main_results}. We report the average results of the evaluated benchmarks here.

	\begin{spacing}{1.5}
	\end{spacing}
	\noindent
	\textbf{Diverse retrieval strategies.}\quad
	The following strategies are explored in the subsequent analysis. (1) Nearest neighbor (\textit{NN}), the strategy used in the main experiments, retrieves $k$ nearest neighbors of the current sample. (2) Nearest neighbor with BM25 filter (\textit{NN, BM}), where we apply BM25 scoring to filters out NN examples not passing a predefined threshold. Samples with no satisfied examples are used with the vanilla instruction template. (3) Diverse nearest neighbor (\textit{DNN}), retrieves $K$ nearest neighbors with $K>>k$ and randomly selects $k$ examples from them. (4) Diverse nearest with BM25 filter (\textit{DNN,BM}), filters out DNN examples not reaching the BM25 threshold. (5) \textit{Random}, uniformly selects $k$ random examples. (6) Mixed nearest neighbors (\textit{MixedNN}), mixes the using of the NN and random retrieval strategies with the ratio of NN set to $a$.
 
	\begin{figure}[t]
		\centerline{\includegraphics[trim=0 15 0 5, clip,width=\linewidth]{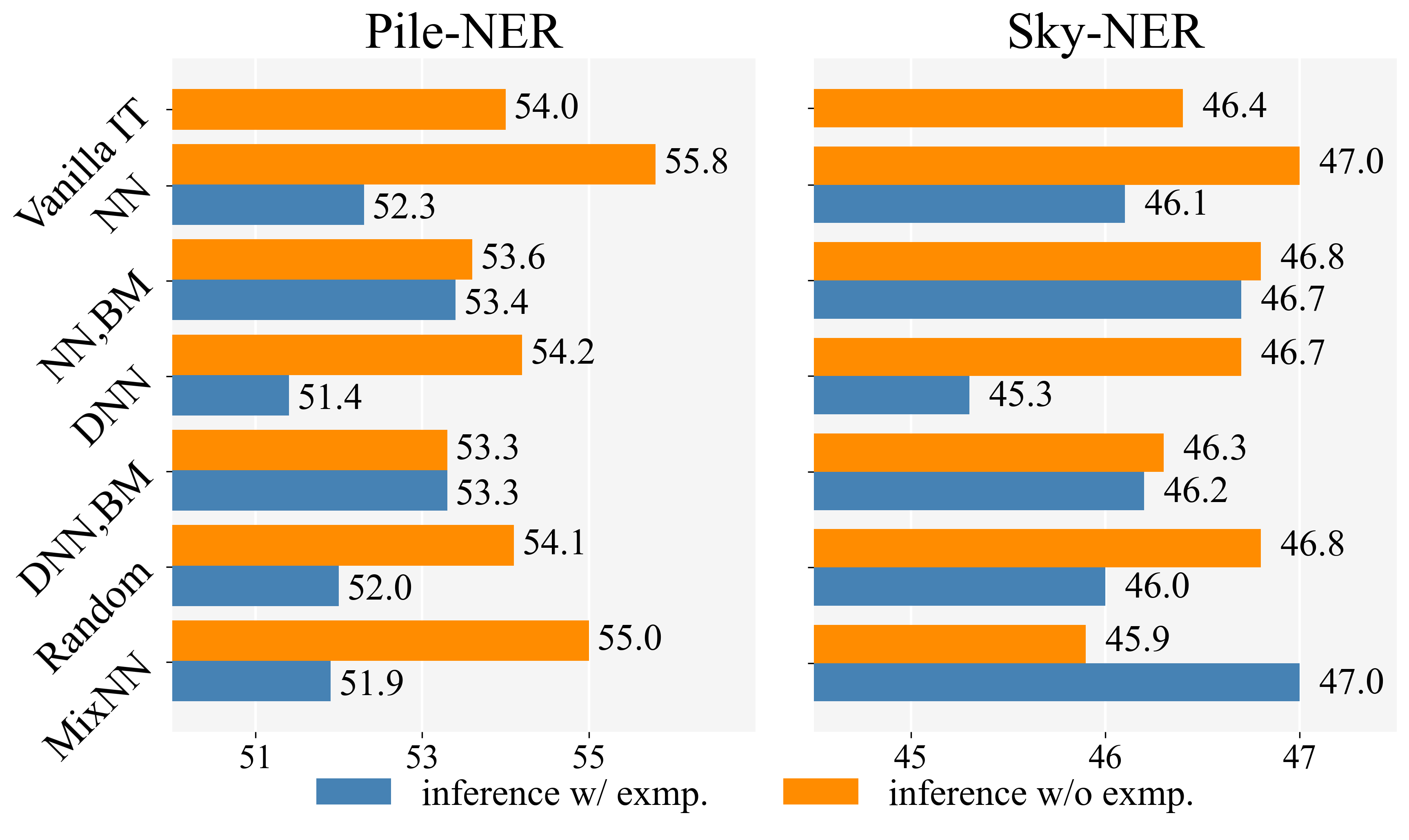}}
		\caption{Impacts of training using various retrieval strategies in RA-IT. The average F1 value of the evaluated benchmarks is reported. \textit{NN} exhibits the best performances, suggesting the need of training with retrieved context.}
		\label{fig:retrieval_strategy_train}
	\end{figure}

	\begin{figure}[t]
		\centerline{\includegraphics[trim=0 20 0 5, clip, width=\linewidth]{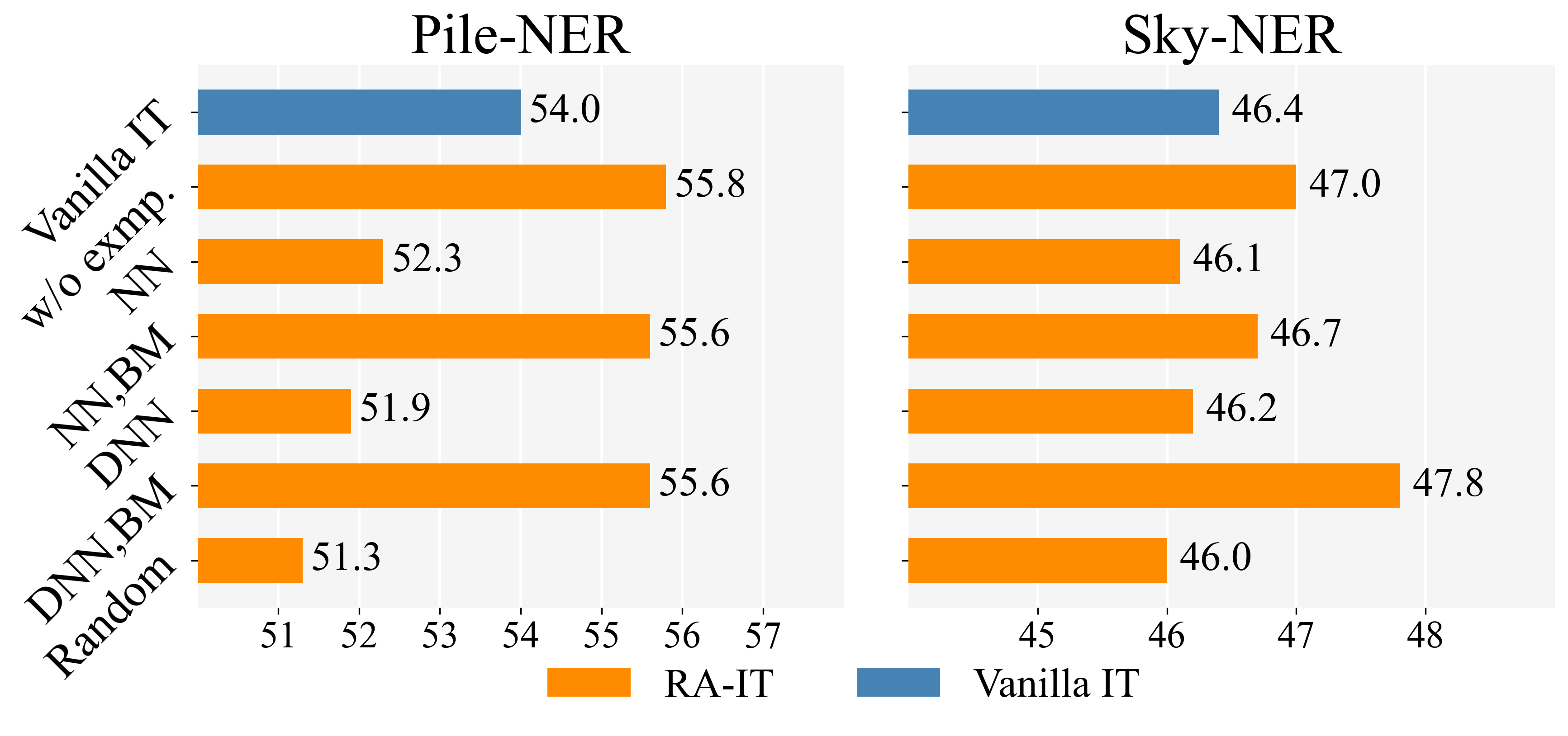}}
		\caption{Impacts of inferece with \textit{out-domain examples} using various retrieval strategies. The average F1 value of the evaluated benchmarks are reported.  \textbf{w/o exmp.} means inference without example. Applying example filtering strategy such as BM25 filtering benefits RAG with out-domain examples.}
		\label{fig:retrieval_strategy_inference}
		\vspace{-0.5em}
	\end{figure}
	
	\begin{figure}[t]
		\centerline{\includegraphics[trim=0 20 0 5, clip, width=\linewidth]{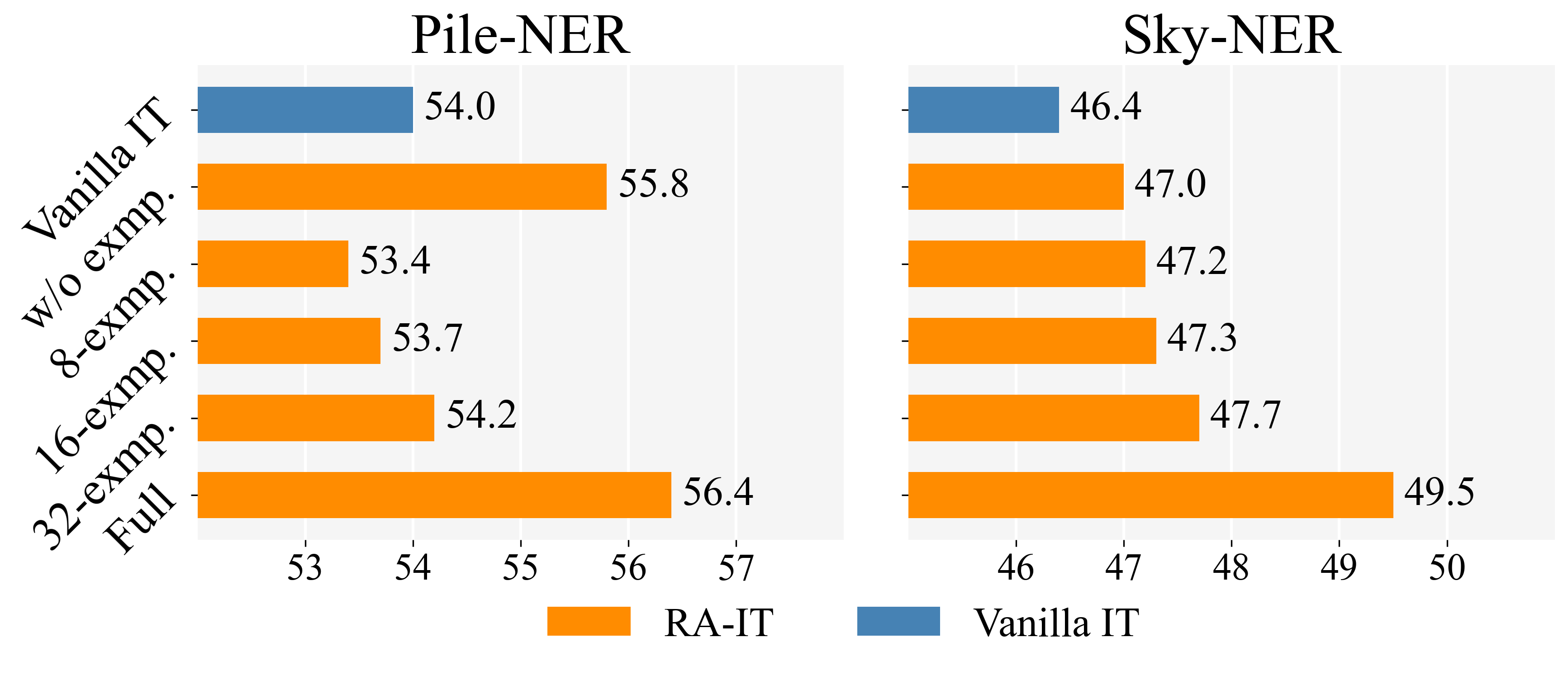}}
		\caption{Impacts of inference with \textit{in-domain examples} using \textit{NN} retrieval. The average F1 value of the evaluated benchmarks are reported. \textbf{$N$-exmp.} means the example pool of size $N$. The results indicate that sufficient in-domain examples are helpful for inference with RAG.}
		\label{fig:fs_5k_zh_en}
	\end{figure}

	\begin{spacing}{1.5}
	\end{spacing}
	\noindent
	\textbf{Training with diverse retrieval strategies}.\quad Fig. \ref{fig:retrieval_strategy_train} visualize the results of training with various retrieval strategies. We conduct inference with and without examples for each strategy, and set the retrieval strategy of inference the same as of training. The most straight forward method \textit{NN} shows best performances, suggesting the benefits of semantically similar examples. \textit{Random} strategy, though inferior to \textit{NN}, also shows improvements, indicating that random examples might introduce some general information of NER taks to the model. Meanwhile, inference with examples does not guarantee improvements and often hurt performances. This may due to the differences of the annotation schema between the automatically constructed data and the human-annotated benchmarks.

	\begin{spacing}{1.5}
	\end{spacing}
	\noindent
	\textbf{Inference with out-domain examples}.\quad
	During inference, since examples from the automatically constructed data is not aligned with the domains and schemas of the human-annotated benchmarks, we refer to them as \textit{out-domain examples}. Fig. \ref{fig:retrieval_strategy_inference} shows the results of inference with out-domain examples using diverse retrieval strategies. We use the model trained with NN strategy here. After applying example filtering such as BM25 scoring, inference with out-domain examples shows improvements compared to the baseline, suggesting the need of example filtering when implementing RAG with out-domain examples.

	\begin{spacing}{1.5}
	\end{spacing}
	\noindent
	\textbf{Inference with in-domain examples}.\quad
	We explore the setting where a few \textit{in-domain examples} are available for inference. We randomly sample an example pool of size $N$ from the original training sets of the benchmarks, then retrieve $k$ NN from this pool as in-domain examples. We also evaluate on \textit{full} pool where the entire training set is used for retrieval. Results are shown in Fig. \ref{fig:fs_5k_zh_en}. In-domain examples show substantial improvements in Chinese. Meanwhile, sufficient in-domain examples are required for improvements in English. This indicates the benefits of providing sufficient in-domain examples for RAG.
	
	Based on the above analysis, we suggest implementing on-demand RAG for inference after RA-IT. When sufficient in-domain examples are available, conduct RAG with similar examples to boost inference. When only out-domain examples are available, apply an example filtering method such as BM25 scoring for RAG, or simply conduct inference without examples.

\section{Related Work}
	\subsection{IE with LLMs}
	The main techniques studied in the area of IE with LLMs fall under advanced prompt designing\citep{guo2023retrievalaugmented,xie2023empirical,wang2023gptner}, instruction tuning (IT) \citep{sainz2024gollie,zhou2024universalner,li2024knowcoder} and data augmentation \citep{josifoski2023exploiting,zhang2023llmaaa,ma2023star}. Many of the prompt designing methods apply RAG to an off-the-shelf LLM to assist inference \citep{guo2023retrievalaugmented,wan2023gptre,xie2024selfimproving}, which retrieves similar examples to provide more useful information for the LLM. Works of IT incorporate the information for IE into the LLMs through task-specific fine-tuning \citep{sainz2024gollie,zhou2024universalner}.
	Different from previous works, we explore retrieval augmented IT (RA-IT) for IE, with a focus on the open NER task. 
	
	Following UniNER \citep{zhou2024universalner}, we conduct investigations under the targeted distillation setting, since UniNER successfully distills the strong capability of ChatGPT in open NER into a smaller model without any human-annotated data. Other works of IT for IE, \citet{sainz2024gollie,li2024knowcoder} adopt the code-style instruction to fine-tune LLMs in effectively generating IE outputs through code generation. They are \textbf{orthogonal} to this work since the strategy of RA-IT can be integrated into various styles of instructions. Moreover, \citet{zaratiana2023glinergeneralistmodelnamed} integrated the strong capability of ChatGPT in open NER into smaller-scale bidirectional LMs (BiLMs) such as BERT \citep{devlin-etal-2019-bert}. How to integrate retrieval augmentation into the BiLMs frameworks is also worth exploring in future work.
	
	
	
	\subsection{Retrieval aware Fine-Tuning}
	Retrieval augmented generation (RAG) has achieved large improvements in diverse tasks with the off-the-shelf LLMs \citep{ram-etal-2023-context}. Recent works has explored retrieval aware IT for LLMs \citep{jiang-etal-2023-active,zhang2024raft}. \citet{jiang-etal-2023-active} pre-trains a retriever and LM jointly, then conducts few-shot fine-tuning on downstream tasks. \citet{luo2023sail} instruction-tunes LMs with retrieved passages prepended to inputs. \citet{zhang2024raft} retrieves both gold and distractor documents for IT to make the model resistant to unhelpful documents. \citet{liu2024chatqa} explores context-enhanced IT to enhance model's capability for conversational QA over a given context. However, retrieval augmented and context-enhanced IT has remained unexplored in IE. We fill this gap and explore (RA-IT) on the task of open domain NER.
	
	

	\section{Conclusion}
	This paper explores RA-IT for open NER. We retrieve semantically similar examples to form the context-enhanced instruction data. RA-IT achieves consistent improvements across various data sizes in English and Chinese, suggesting the need of context-enhanced training. Thorough analysis verifies the benefits of semantically similar examples for training and the need of example filtering and in-domain examples for inference.

    \section*{Limitations}
    This work faces the following limitations: 
 
    (1) Although the RA-IT strategy improves the open NER performance, it does not guarantee improvements when using RAG during inference. Applying some example filtering strategies and introducing in-domain examples alleviate this problem, but the effectiveness is till marginal. More advanced approaches of improving RA-IT models in conducting RAG for open NER are worth exploring. 
    
    (2) The investigation of data efficiency in this work is merely a small preliminary empirical study. However, data efficiency, such as selecting most influential and beneficial data is important for real-world applications of IE since it might effectively save computation and annotation costs.

     \section*{Acknowledgements}
    This research is supported by Zhejiang Provincial Natural Science Foundation of China (LDT23F02023F02) and the National Natural Science Foundation of China (72350710798).

	\bibliography{custom}
	
	\newpage
	
	\appendix

\begin{table*}[t]
		\centering
		\small
		\begin{tabular}{lcccccccccc}
			\toprule
			\multicolumn{2}{c}{Model}  & Ontonotes 4 & MSRA & Weibo & Boson & ClueNER & CMeEE & Ren. & Yidu & Avg. \\
			\midrule
			\multirow{2}{*}{ChatGPT} & Teacher & 29.7 & 41.4 & 30.3 & 46.7 & 44.8 & 43.2 & 34.3 & 34.9 & 38.1 \\
			& Student & 48.0 & 52.6 & 38.0 & 52.2 & 42.4 & 41.0 & 48.8 & 49.0 & \textbf{46.5} \\
			\midrule
			\multirow{2}{*}{Claude} & Teacher & 35.0 & 46.4 & 29.9 & 35.7 & 46.3 & 43.1 & 34.1 & 31.1 & 37.7  \\
			& Student & 50.1 & 52.6 & 36.9 & 47.2 & 44.0 & 42.6 & 49.7 & 46.6 & \textbf{46.2} \\
			\midrule
			\multirow{2}{*}{Moonshot} & Teacher & 43.4 & 51.6 & 27.5 & 51.6 & 49.6 & 43.1 & 41.8 & 34.7 & 42.9 \\
			& Student & 50.8 & 49.7 & 34.9 & 54.3 & 43.5 & 43.0 & 48.2 & 52.1 & \textbf{47.0} \\
			\midrule
			\multirow{2}{*}{GLM-4} & Teacher & 36.7 & 49.4 & 28.3 & 38.3 & 49.1 & 45.9 & 34.4 & 35.5 & 39.7 \\
			& Student & 47.9 & 45.2 & 35.0 & 49.1 & 42.8 & 44.1 & 45.2 & 48.1 & \textbf{44.7} \\
			\bottomrule
		\end{tabular}
		\caption{The method of LLMs' extraction and LLM-guided SFT on Chinese dataset. "Teacher" refers to the results directly returned by instruct-LLM. "Student" refers to the results after SFT using the UniNER method by LLMs' results.}
		\label{tab:chinese_diverse_teacher}
	\end{table*}

    \section{Chinese Data Construction}
	\label{sec:appendix_zh_data_construction_prompt}

\begin{figure}[h]
		\centerline{\includegraphics[width=0.9\linewidth]{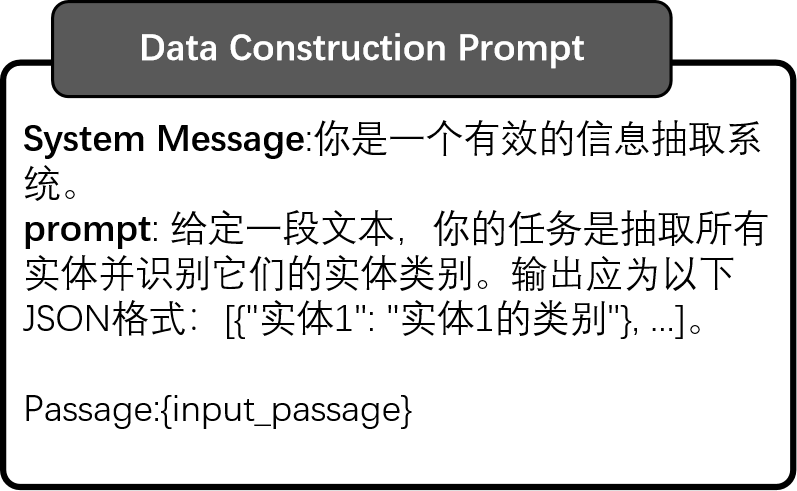}}
		\caption{Data construction prompt for Chinese open domain NER.}
		\label{fig:zh_data_construction_prompt}
	\end{figure}

	\begin{spacing}{1.5}
	\end{spacing}
	\noindent
	\textbf{Data construction prompt.}\quad
	Fig. \ref{fig:zh_data_construction_prompt} shows the prompt used for Chinese distillation data construction. We follow \citet{zhou2024universalner} to design the prompt for Chinese data construction. We adopt the data construction prompt of Pile-NER-type \footnote{https://huggingface.co/datasets/Universal-NER/Pile-NER-type}, since it shows the best performance as in \citep{zhou2024universalner}.

	\begin{spacing}{1.5}
	\end{spacing}
	\noindent
	\textbf{Data processing.}\quad
	Following \citep{zhou2024universalner}, we chunk the passages sampled from the Sky corpus\footnote{https://huggingface.co/datasets/Skywork/SkyPile-150B} to texts of a max length of 256 tokens and randomly sample 50K passages. Due to limited computation resources, we sample the first twenty files in Sky corpus for data construction, since the size of the entire Sky corpus is beyond the processing capability of our machines. We conduct the same data processing procedures including output filtering and negative sampling as in UniNER. Specifically, the negative sampling strategy for entity types, is applied with a probability  proportional to the frequency of entity types in the entire constructed dataset.
	
	\begin{spacing}{1.5}
	\end{spacing}
	\noindent
	\textbf{Instruction data construction.}\quad The instruction tuning data for Chinese scenario is as shown in Table. \ref{tab:Chinese instruct sample}.

	\begin{table*}[t]
		\centering
		\begin{tabular}{l|c|c|ccc}
			\toprule
			Language & Dataset & Labels & Train & Valid & Test \\
			\midrule
			\multirow{7}{*}{English} 
			& CrossNER\_AI & 13 & 100 & 350 & 431 \\
			& CrossNER\_literature & 11 & 100 & 400 & 416 \\
			& CrossNER\_music & 12 & 100 & 380 & 465 \\
			& CrossNER\_politics & 8 & 199 & 540 & 650 \\
			& CrossNER\_science & 16 & 200 & 450 & 543  \\
			& MIT Moive Review & 12 & 9774 & 2442 & 2442 \\
			& MIT Restaurant Review & 8 & 7659 & 1520 & 1520 \\
			\midrule
			\multirow{8}{*}{Chinese} 
			& Ontonotes4 & 4 & 15724 & 4301 & 4346 \\
			& MSRA & 3 & 46364 & - & 4365  \\
			& Weibo & 4 & 1350 & 270 & 270 \\
			& Boson & 6 & 1637 & 184 & 179 \\
			& ClueNER & 10 & 10748 & 1343 & - \\
			& CMeEE & 9 & 15000 & 5000 & - \\
			& Ren. & 4 & 228616 & 28768 & 28885 \\
			& Yidu & 6& 1000 & - & 379 \\
			\bottomrule
		\end{tabular}
		\caption{Dataset statistic.}
		\label{tab:bench_stat}
	\end{table*}

    \section{Diverse Teachers for Data Construction}
	\label{sec:appendix_diverse_teacher}
	
	We explored the effect of using diverse teachers for data construction. We also tried ensemble distillation: multiple teachers are used to annotate entities simultaneously, and the final annotation is acquired by majority voting on the answers from all teachers. The results are shown in Table \ref{tab:chinese_diverse_teacher}
	
	We investigated four representative powerful teacher models, ChatGPT (gpt-3.5-turbo-0125) \citep{openai_blog_22}, Claude 3 (claude-3-haiku) \citep{claude3haiku_blog}, Moonshot (moonshot-v1-8k) \citep{moonshot_platform} and GLM-4 (glm-4) \citep{glm4_blog} for distilling open NER capability.

	\section{More Details of Experimental Settings}
	\label{appendix:exp_setting}
    \subsection{Preliminary Study on Data Efficiency}
    We explore the impact of various data sizes: [0.5K, 1K, 5K, 10K, 20K, 30K, 40K, 50K]. For each data size, we randomly sample two sets of data and report the average. 
    
	\subsection{Training}
	we use the training data Pile-NER released by \citet{zhou2024universalner}, and we adopt the Pile-NER-type version\footnote{https://huggingface.co/datasets/Universal-NER/Pile-NER-type}, which shows better performance than Pile-NER-definition\footnote{https://huggingface.co/datasets/Universal-NER/Pile-NER-definition}. In our practice, we filter out samples with unparseable entity outputs in Pile-NER-type, which finally leaves 45K samples for actual experiments.  
 
    Through our experiments, we train models for 3 epochs with a batch size of 8 and a learning-rate of 5e-5. A cosine scheduler is adopted. Each experiment is run on one single A100 GPU.
	
	\subsection{Retrieval}

	For diverse nearest strategy, we set $K=128$. For example filtering with BM25 scoring, we set the BM25 score threshold as 20.
	
	\subsection{Benchmarks}
	For English, we adopt benchmarks CrossNER \cite{cross_ner_dataset} and MIT-movie/restaurant \cite{mit_dataset}. For Chinese, we collect eight benchmarks across diverse domains, include Ontonotes 4 \cite{weischedel2011-ontonotes}, MSRA \cite{levow-2006-third-msra}, Weibo \cite{peng-dredze-2015-named-weibo}, ClueNER \cite{xu2020cluener2020-cluener}, CMeEE \cite{zhang-etal-2022-cblue-cmeee}, Yidu-S4k \footnote{http://old.openkg.cn/dataset/yidu-s4k}, Boson and PeopleDaily2014(abbreviated as 'Ren.' in the text and tables) \footnote{People Daily 2014 and Boson datasets are available at https://github.com/hspuppy/hugbert/tree/master/ner\_dataset.} The following are our \textbf{sampling strategies} on evaluation data: For evaluated benchmarks, we sample 2000 examples for each test set for evaluation and keep the original test set with fewer than or slightly more than 2000 examples. For those datasets with only the training set and validation set publicly accessible, we randomly select half of the validation data as the test data and the other half as the new validation data.

    \begin{table*}[t]
		\centering
		\small
		\begin{tabular}{lrccccccccc}
			\toprule
			& Data Size & Movie & Restaurant & AI & Literature & Music & Politics & Science & Avg. \\
			\midrule
			ChatGPT & - & 68.50 & 67.00 & 5.30 & 32.80 & 52.40 & 39.80 & 66.60 & 47.50  \\
			\midrule
			\multirow{7}{*}{RA-IT} & 0.5K & 50.92 & 41.50 & 47.61 & 54.27 & 54.15 & 54.78 & 52.16 & 50.77  \\
			& 1K & 44.88 & 39.74 & 50.21 & 56.14 & 55.97 & 56.28 & 54.01 & 51.03 \\
			& 5K & 44.87 & 42.72 & 52.87 & 59.00 & 60.47 & 59.35 & 58.36 & 53.95 \\
			& 10K & 49.81 & 41.47 & 53.78 & 60.99 & 63.79 & 60.84 & 61.47 & 56.02 \\
			& 20K & 50.14 & 42.17 & 57.07 & 62.02 & 65.43 & 61.92 & 63.35 & \textbf{57.44} \\
			& 30K & 47.21 & 40.84 & 58.15 & 63.11 & 65.33 & 62.64 & 64.46 & 57.39 \\
			& 40K & 45.89 & 40.34 & 56.34 & 62.48 & 64.71 & 62.12 & 63.58 & 56.49 \\
			\bottomrule
		\end{tabular}
		\caption{Impact of different dataset sizes on model performance in English scenario. Data size indicates the number of sampled data for prompt fine-tuning.}
		\label{tab:english_data-size}
		\vspace{-0.5em}
	\end{table*}

	\begin{table*}[t]
		\centering
		\small
		\begin{tabular}{lrccccccccc}
			\toprule
			& Data Size & Ontonotes 4 & MSRA & Weibo & Boson & ClueNER & CMeEE & Ren. & Yidu & Avg.  \\
			\midrule
			ChatGPT & - & 29.70 & 41.36 & 30.25 & 46.65 & 44.75 & 43.16 & 34.25 & 34.90 & 38.13 \\
			\midrule
			\multirow{7}{*}{RA-IT}& 0.5K & 47.05 & 47.20 & 37.25 & 44.40 & 43.08 & 37.59 & 40.07 & 42.05 & 42.34 \\
            &1K & 43.55 & 46.48 & 42.41 & 47.58 & 42.87 & 38.28 & 41.39 & 45.23 & 43.47 \\
			&5K & 48.88 & 51.47 & 38.95 & 52.47 & 43.54 & 41.50 & 47.51 & 47.23 & 46.44 \\
			&10K & 46.28 & 52.56 & 39.26 & 52.92 & 45.42 & 42.59 & 47.99 & 47.95 & 46.87 \\
			&20K & 44.05 & 51.94 & 35.52 & 55.67 & 42.97 & 42.46 & 48.68 & 48.65 & 46.24 \\
            &30K & 42.75 & 48.46 & 35.44 & 53.49 & 43.04 & 42.75 & 48.43 & 35.44 & \textbf{48.71} \\  
            &40K & 44.19 & 49.22 & 33.83 & 52.84 & 42.93 & 42.59 & 47.90 & 33.83 & 48.06  \\
			\bottomrule
		\end{tabular}
		\caption{Impact of different dataset sizes on model performance in Chinese scenario. Data size indicates the number of sampled data for prompt fine-tuning.}
		\label{tab:chinese_data-size}
		\vspace{-0.5em}
	\end{table*}

 	\begin{table*}[t]
		\centering
		\small
		\begin{tabular}{lcccccccccc}
			\toprule
			& \#Example & Ontonotes 4 & MSRA & Weibo & Boson & ClueNER & CMeEE & Ren. & Yidu & Avg.  \\
			\midrule
			ChatGPT & - & 29.70 & 41.36 & 30.25 & 46.65 & 44.75 & 43.16 & 34.25 & 34.90 & 38.13 \\
			\midrule
			Vanilla IT & - & 48.88 & 51.47 & 38.95 & 52.47 & 43.54 & 41.50 & 47.51 & 47.23 & 46.44 \\
            \cmidrule{2-11}
			\multirow{5}{*}{RA-IT} & 2 & 49.23 & 53.08 & 37.43 & 52.64 & 43.27 & 43.87 & 48.31 & 48.47 & \textbf{47.04} \\
			& 4 & 48.80 & 53.32 & 35.50 & 50.65 & 43.41 & 43.82 & 48.34 & 48.86 & 46.59 \\
			& 6 & 47.27 & 53.42 & 38.17 & 53.77 & 44.09 & 42.90 & 47.67 & 48.51 & 46.98 \\
			& 8 & 52.20 & 50.56 & 40.18 & 50.10 & 43.69 & 39.17 & 44.53 & 49.45 & 46.24 \\
			& 10 & 42.56 & 35.15 & 33.71 & 43.48 & 37.28 & 30.71 & 37.23 & 49.88 & 38.75 \\
			\bottomrule
		\end{tabular}
		\caption{Impact of different number of example on RA-IT performance of Chinese scenario. \#Example column indicates the number of examples used for RA-IT}
		\label{tab:chinese_exam_size}
		\vspace{-0.5em}
	\end{table*}

	\section{Full results on data size study}
	We provide the full results of preliminary study on data sizes. English dataset results are shown in Table \ref{tab:english_data-size}. When the number of training data is less than 10K, the model performance improves significantly with data increasing. However, the results do not improve the performance after the number of data exceeds 20K. When applied to the Chinese datasets, the threshold increases to 30K in Table \ref{tab:chinese_data-size}

	\section{Extended Analysis}
	\subsection{Varying Numbers of Examples}
	We keep the number of examples as 2 through our main experiments. 
	Here we explore the impact of increasing the number of examples. We found that increasing the number of examples does not guarantee improvements. This is presumably because that the entire inputs get lengthy when the number of examples increases. And the very long input sequence is challenging for the 7B model to understand. The results for the Chinese datasets and the English datasets are shown in Table \ref{tab:english_exam_size} and Table \ref{tab:chinese_exam_size}, respectively.
	
	\subsection{Ablation for Mixed NN}
	We conduct ablation experiments by mixing different sampling methods, constructing the training set using NN strategy and random strategy in various proportions. The samples were mixed at four different ratios of 0.2, 0.4, 0.6, and 0.8, respectively. Additionally, we both used two different strategies which inference with and without using examples. The experimental results for the Chinese and English datasets are shown in Table \ref{tab:english-mix-nn} and Table \ref{tab:chinese-mix-nn}, respectively. The NN strategy performed the best overall, while the random sampling strategy did not contribute to the results.

	\section{Examples of Instruction Data}
	We show examples of instruction data for English and Chinese in Table \ref{tab:English Instruct sample} and Table \ref{tab:Chinese instruct sample}, respectively. For each sample, we used different strategies to find the $k$ number of different examples in the training set. These prompts were formatted as "Text: sample text.\verb|\n| Entity: [\{entity text 1: entity type 1\}, \{entity text 2: entity type 2\}]\verb|\n| " for the large model to learn from. After the model learned from these examples, we had the model read the test sample text. Then, we asked a question for each entity category. The model's answers were organized as the information extraction results for the test samples.

\begin{table*}[t]
	\centering
	\small
	\renewcommand\arraystretch{1}
	\begin{tabular}{c|p{9cm}}
		\toprule
		\multicolumn{2}{l}{Case 1}  \\
		\midrule
		Input text
		& What is the theme song to stand by me?  \\
		\midrule
		Ground Truth
		& [\textbf{\{'stand by me': 'title'\}}] \\
		\midrule
		Answer from Vanilla-IT & [\textbf{\{'stand by me': 'song'\}}, \{'theme song': 'song'\}]  \\
		\midrule
		Answer from RA-IT & [\textbf{\{'stand by me': 'title'\}}, \{'theme song': 'song'\}] \\
		\midrule
		\multicolumn{2}{l}{Case 2}  \\
		\midrule
		Input text
		& How many times has matt damon been jason bourne?  \\
		\midrule
		Ground Truth
		& [\textbf{\{jason bourne: character\}}, \{matt damon: actor\}] \\
		\midrule
		Answer from Vanilla-IT & [\{paul greengrass: director\},\textbf{\{jason bourne: title\}}, \{matt damon: actor\}]  \\
		\midrule
		Answer from RA-IT & [\textbf{\{jason bourne: character\}}, \{matt damon: actor\}] \\
		\bottomrule
	\end{tabular}
	\caption{Cases that RA-IT benefits the long-tail entity types. We exclude the top 30\% of frequent types and regard the remaining types as long-tail types. The entities in bold are long-tail types that are misclassified by vanilla-IT and corrected by RA-IT. These two cases are also commonsense-related.}
	\label{tab:case_long_tail}
\end{table*}

\begin{table*}[t]
	\centering
	\small
	\renewcommand\arraystretch{1}
	\begin{tabular}{c|p{10cm}}
		\toprule
		\multicolumn{2}{l}{Case 1}  \\
		\midrule
		Input text
		& Viral TK phosphorylates aciclovir into its monophosphate form , which is subsequently phosphorylated to active aciclovir triphoshate by cellular kinase s, thus selectively inhibiting viral DNA polymerase.  \\
		\midrule
		Ground Truth
		& [\textbf{\{Viral TK: enzyme\}, \{cellular kinase: enzyme\}}, \{DNA polymerase: enzyme\}, \{aciclovir triphoshate: chemical compound\}] \\
		\midrule
		Answer from Vanilla-IT & [\textbf{\{Viral TK: scientist\}, \{cellular kinase s: scientist\}, ...}] \\
		\midrule
		Answer from RA-IT & [\textbf{\{Viral TK: scientist\}, \{cellular kinase s: scientist\}, ...}] \\
		\midrule
		\multicolumn{2}{l}{Case 2}  \\
		\midrule
		Input text
		& NIST also differs from Bilingual evaluation understudy in its calculation of the brevity penalty insofar as small variations in translation length do not impact the overall score as much.  \\
		\midrule
		Ground Truth
		& [\textbf{\{NIST: metrics\}, \{bilingual evaluation understudy: metrics\}}] \\
		\midrule
		Answer from Vanilla-IT & [\textbf{\{NIST: organization\}, \{bilingual evaluation understudy: organization\}, ...}] \\
		\midrule
		Answer from RA-IT & [\textbf{\{NIST: organization\}, \{bilingual evaluation understudy: organization\}, ...}] \\
		\bottomrule
	\end{tabular}
	\caption{Cases where RA-IT fails to improve. The entities in bold are vanilla-IT wrongly recognized, and RA-IT failed to improve. These professional entities in biomedical or AI domains require domain knowledge to be recognized.}
	\label{tab:case_knowledge}
\end{table*}

    \section{Case Study}
    We conduct case study to explore the advantages and disadvantages of the proposed RA-IT method. Table \ref{tab:case_long_tail} are two cases that demonstrate that RA-IT benefits the long-tail entity types. We exclude the top 30\% of frequent types and regard the remaining types as long-tail types. The entities in bold are long-tail types that are misclassified by vanilla-IT and corrected by RA-IT. These two cases are also commonsense-related.
    
    Table \ref{tab:case_knowledge} are some bad cases where RA-IT fails to improve. The entities in bold are vanilla-IT wrongly recognized, and RA-IT failed to improve. These professional entities in biomedical or AI domains require domain knowledge to be recognized. This shows that RA-IT benefits commonsense-related cases more than knowledge-seeking cases.

	 	\begin{table*}[t]
		\centering
		\small
		\begin{tabular}{lcccccccccc}
			\toprule
			Method & \#Example & Movie & Restaurant & AI & Literature & Music & Politics & Science & Avg. \\
			\midrule
			ChatGPT & - & 68.50 & 67.00 & 5.30 & 32.80 & 52.40 & 39.80 & 66.60 & 47.50  \\
			\midrule
			Vanilla-IT & - & 52.87 & 59.00 & 60.47 & 59.35 & 58.36 & 44.87 & 42.72 & 53.95 \\
            \cmidrule{2-10}
			\multirow{5}{*}{RA-IT} & 2 & 52.61 & 60.01 & 63.04 & 60.02 & 58.91 & 50.26 & 45.75 & \textbf{55.80} \\
			 & 4 & 51.08 & 59.30 & 62.40 & 59.18 & 58.15 & 51.15 & 45.88 & 55.31 \\
			 & 6 & 48.79 & 54.46 & 55.75 & 54.62 & 50.93 & 54.79 & 46.81 & 52.31 \\
			 & 8 & 34.61 & 36.17 & 34.12 & 34.30 & 35.27 & 45.29 & 36.50 & 36.61 \\
			 & 10 & 22.89 & 30.35 & 22.03 & 30.24 & 26.41 & 38.31 & 35.80 & 29.43 \\
			\bottomrule
		\end{tabular}
		\caption{Impact of different number of example on RA-IT performance of English scenario. \#Example column indicates the number of examples used for RA-IT}
		\label{tab:english_exam_size}
		\vspace{-0.5em}
	\end{table*}

 \begin{table*}[t]
    \centering
    \small
    \begin{tabular}{lccccccccccc}
        \toprule
       Method & ratio & \#Exam. & Movie & Restaurant & AI & Literature & Music & Politics & Science & Avg. \\
        \midrule
        ChatGPT & - & - & 68.50 & 67.00 & 5.30 & 32.80 & 52.40 & 39.80 & 66.60 & 47.49\\
        \midrule
        Vanilla-IT & - & & 44.87 & 42.72 & 52.87 & 59.00 & 60.47 & 59.35 & 58.36 & 53.95\\
        \midrule
        \multirow{12}{*}{RA-IT} & \multirow{2}{*}{0.2} & 0 & 45.65 & 45.25 & 51.89 & 58.36 & 62.00 & 59.26 & 58.21 & 54.37\\
         & & 2 & 41.58 & 39.04 & 50.21 & 57.21 & 59.23 & 58.89 & 58.30 & 52.31\\
         \cmidrule{2-11}
         & \multirow{2}{*}{0.4} & 0 & 46.81 & 45.62 & 51.72 & 58.92 & 62.42 & 59.39 & 58.62 & 54.78\\
         & & 2 & 42.84 & 39.24 & 50.05 & 57.48 & 59.85 & 59.16 & 58.37 & 52.43\\
         \cmidrule{2-11}
         & \multirow{2}{*}{0.6} & 0 & 48.56 & 45.05 & 51.5 & 58.89 & 61.47 & 59.05 & 57.18 & 54.53\\
         & & 2 & 42.92 & 37.70 & 49.99 & 57.41 & 59.01 & 58.76 & 57.84 & 51.95\\
         \cmidrule{2-11}
         & \multirow{2}{*}{0.8} & 0 & 47.82 & 45.49 & 52.29 & 58.81 & 61.90 & 59.61 & 58.70 & 54.95\\
         & & 2 & 41.60 & 37.72 & 49.89 & 57.63 & 59.62 & 58.89 & 57.69 & 51.86\\
         \cmidrule{2-11}
         & \multirow{2}{*}{1} & 0 & 53.20 & 47.36 & 52.50 & 60.86 & 63.13 & 60.77 & 61.02 & \textbf{56.98}\\
         & & 2 & 43.24 & 38.56 & 49.71 & 58.29 & 60.74 & 59.94 & 59.10 & 52.80\\
        \bottomrule
    \end{tabular}
    \caption{Results of mixing random strategy with NN strategy at different ratios in English dataset. 'ratio' indicates the proportion of NN strategy in the total number of samples while training. When ratio=1, all samples are from NN strategy. '\#exam.' indicates whether example data was added to prompts during testing, with 0 indicating no addition, and 2 indicating 2 examples retrieved by NN strategy added to the test examples.}
    \label{tab:english-mix-nn}
    \vspace{-0.5em}
\end{table*}

\begin{table*}[t]
    \centering
    \small
    \begin{tabular}{lccccccccccc}
        \toprule
         Method & ratio & \#Exam. & Ontonotes 4 & MSRA & Weibo & Boson & ClueNER & CMeEE & Ren. & Yidu & Avg.  \\
        \midrule
        ChatGPT  & - & - & 29.70 & 41.36 & 30.25 & 46.65 & 44.75 & 43.16 & 34.25 & 34.90 & 38.13 \\
        \midrule
        Vanilla-IT & - & - & 48.88 & 51.47 & 38.95 & 52.47 & 43.54 & 41.50 & 47.51 & 47.23 & 46.44 \\
        \midrule
        \multirow{12}{*}{RA-IT} & \multirow{2}{*}{0.2} & 0 & 49.36 & 52.53 & 37.57 & 52.42 & 42.86 & 43.44 & 48.45 & 48.03 & 46.83\\
         &  & 2 & 47.93 & 49.97 & 37.30 & 51.40 & 43.10 & 41.71 & 47.97 & 47.35 & 45.84\\
         \cmidrule{2-12}
         & \multirow{2}{*}{0.4} & 0 & 49.03 & 52.71 & 37.81 & 52.52 & 42.83 & 43.53 & 48.33 & 48.04 & 46.85\\
          & & 2 & 48.03 & 50.30 & 37.33 & 51.43 & 43.17 & 41.59 & 47.62 & 46.99 & 45.81\\
          \cmidrule{2-12}
         & \multirow{2}{*}{0.6} & 0 & 48.91 & 52.43 & 37.01 & 51.35 & 42.59 & 43.68 & 48.24 & 48.04 & 46.53\\
         &  & 2 & 47.83 & 49.85 & 37.62 & 53.03 & 42.92 & 41.77 & 47.79 & 47.87 & 46.09\\
         \cmidrule{2-12}
         & \multirow{2}{*}{0.8} & 0 & 49.25 & 52.86 & 37.27 & 52.88 & 43.15 & 43.92 & 48.32 & 48.35 & 47.00\\
         &  & 2 & 48.04 & 49.95 & 38.12 & 50.61 & 43.04 & 41.58 & 47.81 & 47.63 & 45.85\\
         \cmidrule{2-12}
         & \multirow{2}{*}{1} & 0 & 50.46 & 53.73 & 38.10 & 54.36 & 43.70 & 43.78 & 48.65 & 48.87 & \textbf{47.71}\\
         & & 2 & 47.96 & 50.90 & 40.00 & 53.68 & 43.98 & 42.09 & 47.75 & 48.56 & 46.87\\
        \bottomrule
    \end{tabular}
    \caption{Results of mixing random strategy with NN strategy at different ratios in Chinese dataset. 'ratio' indicates the proportion of NN strategy in the total number of samples while training. When ratio=1, all samples are from NN strategy. '\#exam.' indicates whether example data was added to prompts during testing, with 0 indicating no addition, and 2 indicating 2 examples retrieved by NN strategy added to the test examples.}
    \label{tab:chinese-mix-nn}
    \vspace{-0.5em}
\end{table*}
	
	\begin{table*}[t]
		\centering
		\small
		\begin{tabular}{c|p{12cm}}
			\toprule
			\multicolumn{1}{c|}{Role} & \multicolumn{1}{c}{Conversation} \\
			\midrule
			\multirow{40}{*}{Human} 
			& Here are some examples of named entity recognition: \newline Text: 50 Top B2B Marketing Influencers 2017. It's October and you know what that means? Its B2B Marketing influencer speaker list time again. One of my all-time favorite conferences is MarketingProfs B2B Forum in Boston and for the past few years. I've had some fun listing out a top list of speakers ranked by influence around the topic of "B2B marketing". As usual, I used the influencer marketing platform Traackr to import the list of speakers from \#mpb2b 2017 and rank them according to a combination of topical resonance and relevance as well as network reach related to "b2b marketing". Of course, use of their platform in this way is like 1\% of what Traackr can do. I imagine they cringe every time I use their robust tool for such a simple list - but hey, they provide me with access and I use the tool as I see fit. To clarify, my agency TopRank Marketing is also a paying customer of the Traackr platform for clients, where it is used in support of B2B influencer marketing programs for brands like SAP, BMC Software, McKesson and others in ways that are more in line with the platform's capabilities. This is a legit list that recognizes people creating content around B2B marketing that resonates with their social following. \newline Entity: [ \{'TopRank Marketing': 'agency'\}, ... \{'mpb2b 2017': 'event'\} ] \newline Text: How to know and choose online games: differences between current and potential players. This study investigated how different adolescent players acquire game information and the criteria they use in choosing online games and found that (1) current players generally use comprehensive information sources more than potential players do; (2) current players rely on free trials and smooth display of motion graphics as choice criteria more than potential players do; (3) potential players rely on the look of advertisements more than current players do; (4) both current and potential players most likely use word-of-mouth and gaming programs on TV as information sources; and (5) endorser attractiveness is ranked the least important among six choice criteria by both current and potential players. \newline Entity: [\{'online games': 'Product'\}, ... \{'potential players': 'Person'\}] \\
			\midrule
			LLM
			& I've read these examples. \\
			\midrule
			\multirow{10}{*}{Human} 
			& Text: For brands looking to get into the rising world of esports, sponsoring live streamers on twitch is a popular choice. However, it appears competition may be heating up for endorsements from the top 0.2\% of talent. These superstars typically represent the top 50 channels online at any given time and typically have lined up between 2 and 8 simultaneous sponsors leading to a crowded space for brands looking to get involved with esports. Meanwhile, brands able to engage popular streamers below the superstar level, are on average the channel's only sponsor. At Endorse.gg we provide an integrated analytics and engagement platform to help manage large scale and largely exclusive campaigns across a much greater number of these smaller but high quality broadcasters. Source: Analysis of live streamers on twitch.tv over the course of a week. "Top 10 channels" refers to the top 10 online at any given moment. Based on an analysis of currently live channels at several points over the course of a week, we see that the top 10 channels by number of viewers typically command \~15\% of the total twitch viewership and have \~3 actively endorsed brands. Meanwhile, channels \#11-50 capture an additional 25\% of the views while promoting 2.2 brands on average. \\
			\midrule
			LLM
			& I've read this text. \\
			\midrule
			Human
			& What describes "organization" in the text? \\
			\midrule 
			LLM
			& ["brands", "twitch", "Endorse.gg"] \\
			\bottomrule
		\end{tabular}
		\caption{A English sample of prompts and LLM's responses. In the Role column, "Human" indicates the prompt and "LLM" indicates the LLM response.}
		\label{tab:English Instruct sample}
		\vspace{-0.5em}
	\end{table*}
	
	\begin{CJK}{UTF8}{gbsn}
		\begin{table*}[t]
			\centering
			\small
			\begin{tabular}{c|p{12cm}}
				\toprule
				\multicolumn{1}{c|}{Role} & \multicolumn{1}{c}{Conversation} \\
				\midrule
				\multirow{10}{*}{Human} 
				& 以下是一些命名实体识别的例子：\newline 文本：创刊以来，办刊质量和水平不断提高，主要栏目设置有：论著、社区卫生管理、综述讲座、社区疾病预防、社区健康教育、卫生政策、社区计划生育等。 期刊收录：知网收录(中)上海图书馆馆藏维普收录(中)国家图书馆馆藏万方收录(中) 《四川医学》杂志在全国影响力巨大，创刊于1980年，公开发行的月刊杂志。创刊以来，办刊质量和水平不断提高，主要栏目设置有：基金论文、实验研究、临床研究、医学进展、自学辅导站、卫生管理、检验医学、调查报告等。 期刊收录：知网收录(中)万方收录(中)国家图书馆馆藏维普收录(中)CA 化学文摘(美)上海图书馆馆藏统计源核心期刊(中国科技论文核心期刊) 《中国医学工程》杂志在全国影响力巨大，创刊于2002年，公开发行的月刊杂志。\newline 实体：[{'期刊收录': '事件'}, ... ,{'检验医学': '类别'}] \newline 文本：《中国骨伤》杂志是我国大陆地区第一个被美国《医学索引》（IndexMedic 本刊由国家中医药管理局主管，中国中医科学院中药研究所和中华中医药学会主办，是以报道、介绍中医药实验方剂研究为主旨的专业性学术期刊，创刊于1995年10月。本刊办刊方针以提高为主、提高与普及相结合。设置栏目有中医临床、药理、毒理、药物代谢、药剂与炮制、化学与分析、资源与鉴定、综述、专论、数据挖掘、中医传承、标准与规范等，交流方剂及其主要组成药物的研究成果与最新进展。读者对象是从事中、 《中华中医药杂志》原名《中国医药学报》，是中国科学技术协会主管、中华中医药学会主办的国家级高级中医药学术期刊，是中国科学技术协会所属的自然科技期刊中反映中医药研究进展的中医学科杂志。\newline 实体：[{'国家级高级中医药学术期刊': '期刊'}, ... ,{'毒理': '栏目'}]  \\
				\midrule
				LLM 
				& 我已读完这些例子。 \\
				\midrule
				Human 
				& 文本：《福建医药》是综合性医药学术期刊，综合因子为：0.164，期刊级别为：省级期刊。福建医药及时报道国内外医药卫生科研成果及防治疾病经验，反映其进展及水平，为省内外各级医药卫生技术人员提供学术交流平台。 期刊收录：国家图书馆馆藏上海图书馆馆藏维普收录(中)知网收录(中)万方收录(中)CA 化学文摘(美) 《机械工程与自动化》（CN：14-1319/TH）是一本有较高学术价值的大型双月刊，自创刊以来，选题新奇而不失报道广度，服务大众而不失理论高度。颇受业界和广大读者的关注和好评。 《齐齐哈尔大学学报》是省级期刊、荣获首届全国优秀社科学报，综合影响因子为0.074。 \\
				\midrule
				LLM 
				& 我已读完这段文本。 \\
				\midrule
				Human 
				& 文本中属于"组织"的实体有哪些？ \\
				\midrule 
				LLM 
				& [] \\
				\midrule
				Human 
				& 文本中属于"期刊"的实体有哪些？ \\
				\midrule 
				LLM 
				& ["齐齐哈尔大学学报", "机械工程与自动化", "福建医药"] \\
				\midrule
				Human 
				& 文本中属于"收录机构"的实体有哪些？ \\
				\midrule 
				LLM 
				& ["上海图书馆", "维普", "万方", "CA 化学文摘", "知网", "国家图书馆"] \\
				\bottomrule
			\end{tabular}
			\caption{A Chinese sample of prompts and LLM's responses. In the Role column, "Human" indicates the prompt and "LLM" indicates the LLM response. The format of the Chinese prompts is the same as the English.}
			\label{tab:Chinese instruct sample}
			\vspace{-0.5em}
		\end{table*}
	\end{CJK}

\end{document}